# IMAGE TYPE WATER METER CHARACTER RECOGNITION BASED ON EMBEDDED DSP

LIU Ying[1], HAN Yan-bin[2] and ZHANG Yu-lin[3]

[1]School of Information Science and Engineering, University of Jinan, Jinan 250022, PR China
75587050@qq.com

[2] School of Information Science and Engineering, University of Jinan, Jinan 250022, PR China
87685630@qq.com

[3] School of Information Science and Engineering, University of Jinan, Jinan 250022, PR China
ise_zhangyl@ujn.edu.cn

## ABSTRACT

*In the paper, we combined DSP processor with image processing algorithm and studied the method of water meter character recognition. We collected water meter image through camera at a fixed angle, and the projection method is used to recognize those digital images. The experiment results show that the method can recognize the meter characters accurately and artificial meter reading is replaced by automatic digital recognition, which improves working efficiency.*

## 1. INTRODUCTION

In recent years, with the progress of society and science technology, it is quite common that manual work had been replaced by machines. In our daily life, water consumption statistics were mostly gotten in the form of artificial meter reading. This method not only increased the workload and difficulty of work, but also had some obvious drawbacks. The use of automatic meter reading systems not only overcome the insufficiency of manual meter reading, but also reduced the practical resources and save time.

Based on DSP chip, this study collected water meter image through the camera on hardware system. Firstly, we achieved the purpose of automatic identification through a series of image processing technology. Lastly, the designed algorithm was transplanted into hardware devices and tested. (As shown in figure 1) The advantages of the study are simply structure with high recognition rate and low cost. The identification process is mainly divided into two parts: image processing and water meter characters recognition.

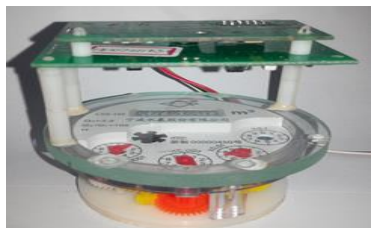
Figure 1 hardware test

## 2. RELATED WORK

Image processing is an important task, and it directly affects the recognition of the digital image. In the techniques of image processing, the main techniques that related to this study include image filtering, image enhancement, edge detection, image segmentation and image binarization, etc.

For example, based on the quality of image, image segmentation is divided into many concrete methods. In literature 1, it segmented digital image of water meter through counting the number of pixels in each row and column. In literature 2, it segmented digital image of water meter through vertical projection and horizontal projection. As for the image binarization, it is divided into fixed threshold method and adaptive threshold method. The literature [2-4] adopted adaptive threshold method. The literature 2 used cv adaptive threshold that was included in Opencv. The literature 3 used OTSU algorithm. And the literature 4 adopted three adaptive threshold methods. In the method of digital recognition, the literature [5, 6] adopted neural network recognition method [7, 8]. The literature 9 adopted template matching method to recognition. And the literature 10 improved template matching and used it to recognize.

In this paper, the methods that we used in image processing included image segmentation, image binarization, and image filtering. In the process of digital meter identification, we adopted the method of projection. Specific implementation process is shown in figure 2.

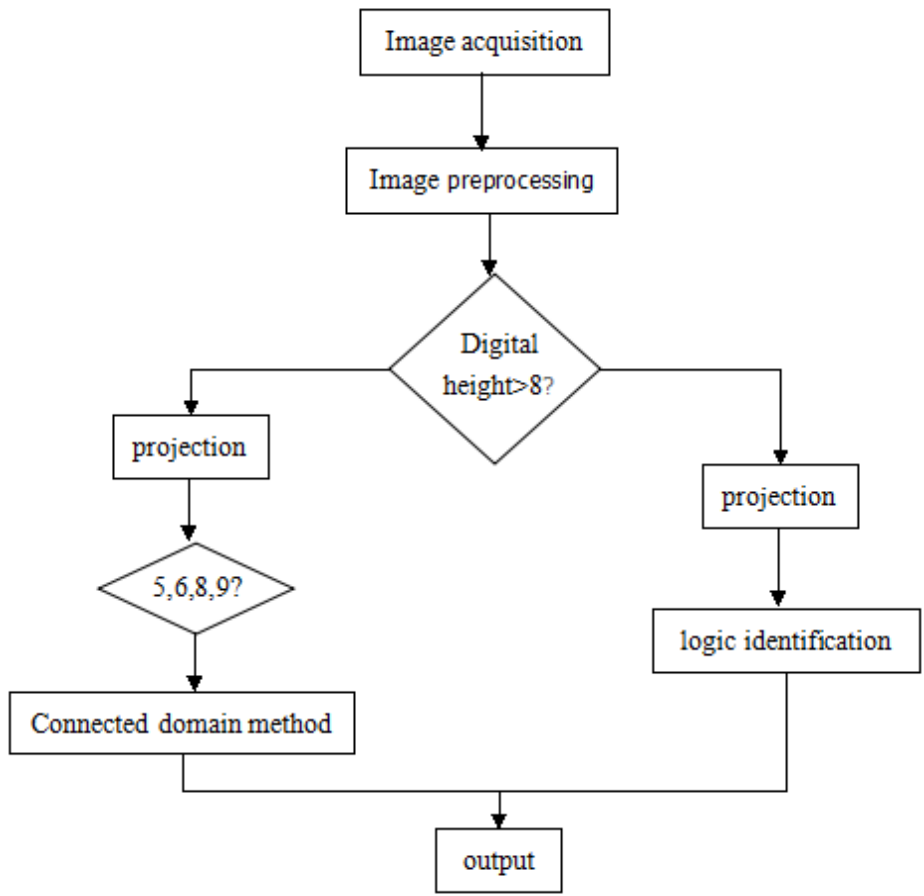

Figure 2 digital meter processing

## 3. WATER METER IMAGE PRE-TREATMENT

Water meter image pre-processing is important to the character recognition. Image pre-treatment can significantly improve the quality of image and reduce noise. It is helpful for the extraction of image characteristics.

### 3.1. Image segmentation

Because the images that were used in this study were collected through the installed cameras, the size of the image was relatively fixed. The collected water meter images had obvious border (shown in figure 3), so we segmented those pictures at a fixed point, and extracted the effective area (shown in figure 4).

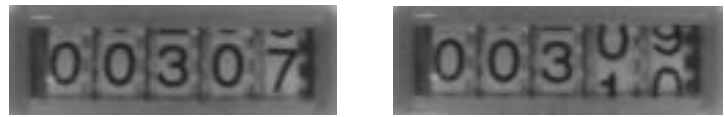

Figure 3 camera collections of pictures

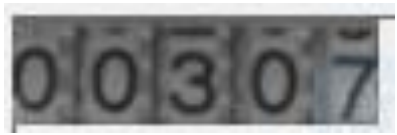

Figure 4 image segmentation

### 3.2. Image binarization

The role of image binarization is to make the image black and white obviously. Binarization method is divided into fixed threshold method and adaptive threshold method. It is because the adaptability of fixed threshold method is poor that we generally use the adaptive threshold method.

In the study, the OTSU algorithm had simple calculation and it was not affected by the image brightness and contrast, so we used OTSU algorithm. After segmenting, the image was divided into five areas. We used the OTSU algorithm in the five areas separately for local binarization. The result is shown in figure 5.

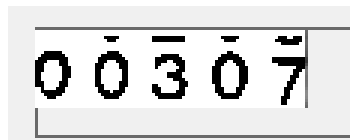

Figure 5 the effect after binarization

### 3.3. Image filtering

Filtering can eliminate the noise points in the image. For binary image, we eliminate the image boundary, and then remove the small area into the isolated pixels. The result is shown in figure 6.

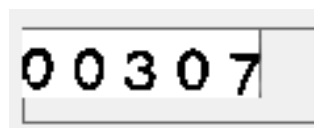

Figure 6 after filtering

# 4. DIGITAL SYMBOL RECOGNITION

There are many methods of digital symbol recognition, such as template matching, neural network recognition, statistical method and fuzzy theory method, etc.

In the research, digital symbol recognition is divided into full-word recognition, half-word recognition and logic recognition. Projection method was used in this paper. This method makes the digital project, and then, matches the projection curve with template curve. Finally, it outputs the number which has the minimum error. Because numbers 5, 6, 8 and 9 have higher similarity in this method, connected domain method was used in the two - times recognition. The connected domain method uses the number of connected domain and the location of connected domain to recognition.

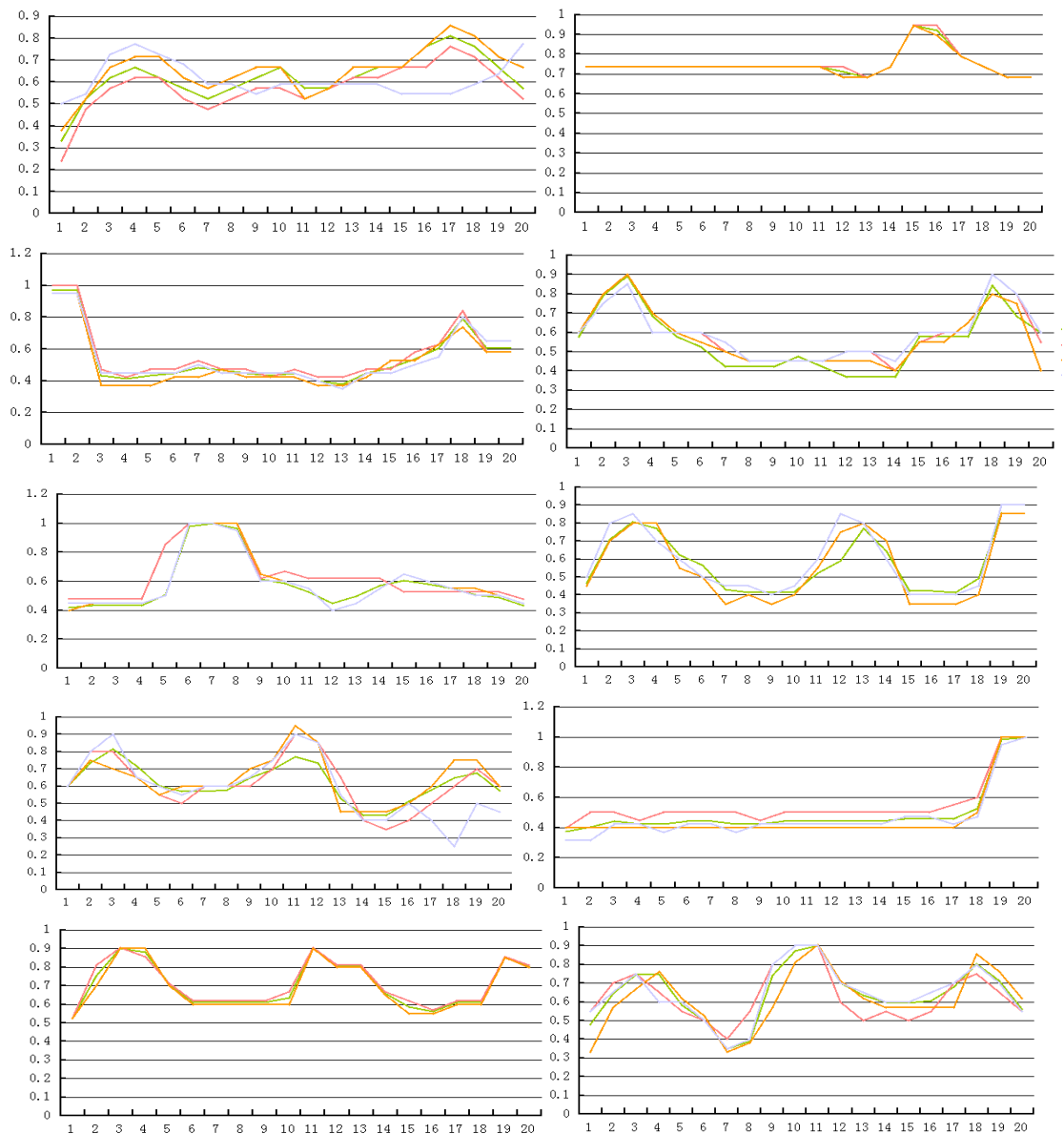

Figure 7 projection graph of 0-9

Figure 7 is the projection curve of 0-9. As for every number, I selected four samples. From figure 7, we can see that the numbers 5, 6, 8 and 9 have similar characteristics. Other digital have obvious characteristics and the projection curve at the same digital has good consistency.

Therefore, projection method has good ability to recognize. As for the problem that the projection method is not applicable in the numbers 5, 6, 8 and 9, we put forward the connected domain method. The method can solve the problem.

## 5. HALF-WORD RECOGNITION

In the above recognition method, the accuracy of full-word recognition is close to 100%. As for the half-word recognition, it was always the difficulty of the study. The common methods are template matching method, and neural network recognition method, etc. Literature 11 was studied specifically for half-word recognition. The half-word is composed of one long and one short, the algorithm is just calculated the longer half-word, and the shorter one was not considered.

In my paper, I improved the template matching method based on the projection method of full-word recognition. (The image of half-word is shown in figure 8) I reserved the long part and the short pat, and then, I used the method of projection on them respectively. Finally, I analysed the matching result and recognized the digital according to the logical continuity.

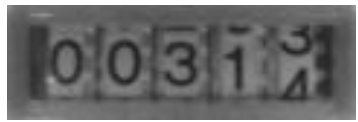

Figure 8 half-word image

Table 1 is the matched data of the half-word in figure 8. In matched data in the above part, the number 3 has the minimum error. In matched data of the below part, the number 4 has the minimum error. Because of the number 3 and the number 4 have continuity in logic, they can recognize correctly.

Table 1 the matched data of the half-word in figure 8

| match object | 0 | 1 | 2 | 3 | 4 | 5 | 6 | 7 | 8 | 9 |
| --- | --- | --- | --- | --- | --- | --- | --- | --- | --- | --- |
| error (above) | 3.5 | 5.8 | 4.4 | 1.6 | 4.9 | 4.0 | 3.0 | 5.4 | 3.6 | 4.1 |
| error (below) | 7.2 | 6.4 | 6.2 | 5.3 | 1.5 | 6.6 | 6.3 | 5.7 | 5.7 | 6.5 |

## 6 LOGIC IDENTIFY

Aiming at several half-word in figure 9, there is needed to adopt logical identification. When the image appears more than one half-word, if we only recognize the longer half-word, there will be a cross recognition (as shown in figure 9, those images may be recognized to 310 and 300) and cause errors. So we need to combine the continuity of digital to recognize. The method not only saves the calculation time, but also improves the recognition accuracy.

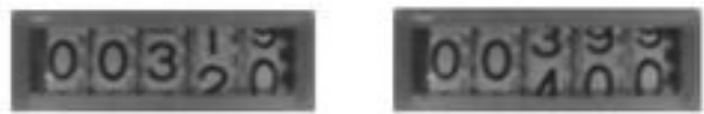

Figure 9 several half-words

The recognition results of some typical half-word are shown in figure 10. The above of the image is original image, the lower left corner is the processed image, and the lower right is recognition results.

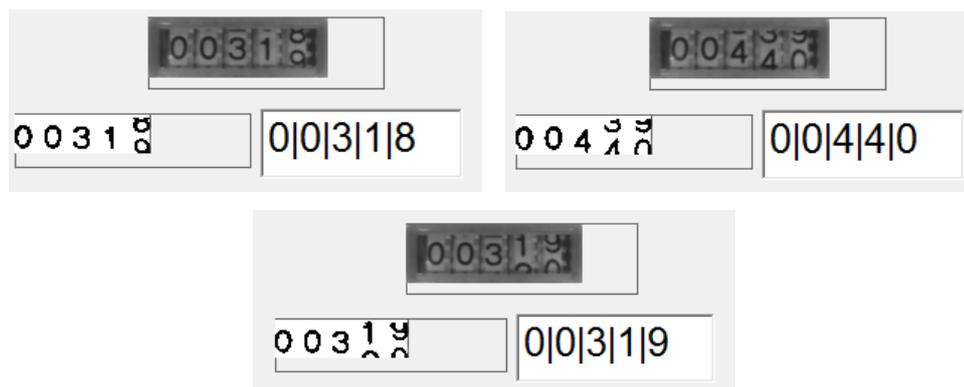

Figure 10 recognition results

## 7 CONCLUSIONS

This study used the projection method as the main means of digital recognition. In addition, we put forward the connected domain method and logical identification method in half-word. In this paper, the scheme of automatic recognition of digital water meter has good results, especially for the difficulty recognition of half-word. What is more, this study adopts the DSP chip. Through combination with embedded technology, making this study has practical application value. Through testing, the method that mentioned in this paper obtained a good accuracy. The recognition accuracy can reach above 95%. The method is simple and effective, and can be used to replace artificial meter reading. It overcomes the disadvantages of manual meter reading and has good practical value.

## ACKNOWLEDGEMENTS

This article is completed under the guidance of professor ZHANG and professor HAN. In addition, my classmates also gave me a lot of help. What is more, University of Jinan provided a good learning environment for us.

**Authors**

LIU Ying (1991- ), female, master, research in switched reluctance motor, pattern recognition, image processing.

HAN Yan-bin (1979- ), male, associate professor, research in pattern recognition, image processing.

ZHANG Yu-lin (1972- ), corresponding author, male, professor, the research direction for embedded system, computer architecture.